%% file: IEEEabrv.tex
\documentclass[journal]{IEEEtran}
\usepackage{times}
\usepackage{epsfig}
\usepackage{graphicx}
\usepackage{amsmath}
\usepackage{amssymb}
\usepackage{overpic}
\usepackage{cite}
\usepackage{enumitem}

\usepackage[ruled]{algorithm2e}

\usepackage{multirow}
\usepackage{booktabs}
\usepackage{color}

\newcommand{\etal}{\emph{et al.}}
\newcommand{\ie}{\emph{i.e.}}

\newcounter{ToDo}
\newcounter{tye_comm}

\definecolor{blue-violet}{rgb}{0.54, 0.17, 0.89}
\definecolor{mygreen}{rgb}{0.0, 0.5, 0.0}
\definecolor{awesome}{rgb}{1.0, 0.13, 0.32}
\definecolor{bostonuniversityred}{rgb}{1.0, 0.0, 0.0}
\ifCLASSINFOpdf
\else
\fi
\begin{document}
\title{OST: Efficient One-stream Network for 3D Single Object Tracking in Point Clouds }
%\title{OST:One-stream Network For 3D Single Object Tracking in Point Clouds }
\author{Xiantong~Zhao,
        Yinan~Han,
        Shengjing~Tian,
        Jian~Liu,
        and~Xiuping~Liu% <-this % stops a space
\thanks{Xiantong~Zhao and Yinan~Han are with the School of Mathematical Sciences, Dalian University of Technology, Dalian, China (e-mail: 934613764@qq.com,601854103@qq.com).}
\thanks{Shengjing~Tian is with the School of Economics and Management, China University of Mining and Technology, Xuzhou, China (e-mail:tye.dut@gmail.com).}
\thanks{Jian~Liu is with School of Software, Tsinghua University, Beijing 100084, China (e-mail: mike2006@mail.tsinghua.edu.cn)}
\thanks{Xiuping Liu is with Dalian University of Technology (e-mail: xpliu@dlut.edu.cn).}
\thanks{Manuscript received --}}


\markboth{Journal of \LaTeX\ Class Files,~Vol.~Null, No.~Null, August~2019}%
{Shell \MakeLowercase{\textit{et al.}}: Bare Demo of IEEEtran.cls for IEEE Journals}



\maketitle

\begin{abstract}
Although recent Siamese network-based trackers have achieved impressive perceptual accuracy for single object tracking in LiDAR point clouds, they advance with some heavy correlation operations on relation modeling and overlook the inherent merit of arbitrariness compared to multiple object tracking. In this work, we propose a radically novel one-stream network with the strength of the Transformer encoding, which avoids the correlation operations occurring in previous Siamese network, thus considerably reducing the computational effort. In particular, the proposed method mainly consists of a Template-aware Transformer Module (TTM) and a Multi-scale Feature Aggregation (MFA) module capable of fusing spatial and semantic information. The TTM stitches the specified template and the search region together and leverages an attention mechanism to establish the information flow, breaking the previous pattern of independent \textit{extraction-and-correlation}. As a result, this module makes it possible to directly generate template-aware features that are suitable for the arbitrary and continuously changing nature of the target, enabling the model to deal with unseen categories. In addition, the MFA is proposed to make spatial and semantic information complementary to each other, which is characterized by reverse directional feature propagation that aggregates information from shallow to deep layers. Extensive experiments on KITTI and nuScenes demonstrate that our method has achieved considerable performance not only for class-specific tracking but also for class-agnostic tracking with less computation and higher efficiency. 

%   In recent years, 3D single object tracking in point clouds has been greatly developed. However, Models trained from one class can not be used in other classes, especially, unknown classes. In this work, we propose a single stream tracker that combines feature learning with feature augmentation. The template will guide the feature extraction of the search area, search region feature extraction adaptive template. This makes the target point features and background point features distinguishable, no matter what class the template belongs to. Extensive evaluation on the KITTI and nuScenes datasets shows that our method significantly outperforms the current state-of-the-art methods by a large margin,specially when tracking unseen objects.   corner case?
  
\end{abstract}


% \fi
\IEEEpeerreviewmaketitle

\input{introduction.tex}

\input{related_work.tex}
\input{proposed_method.tex}

\input{experiments.tex}

\input{conclution.tex}
\ifCLASSOPTIONcaptionsoff
  \newpage
\fi



\bibliographystyle{IEEEtran}
\bibliography{IEEEabrv}
\end{document}

%% file: introduction.tex
\section{Introduction}
\IEEEPARstart{O}{bject} tracking, as an important mean of perceiving the external environment for the agent, has great significance in many fields, such as autonomous driving, robotics vision, visual surveillance \cite{Comport2004RobustMT, Luo2018FastAF, Machida2012HumanMT}. Recently, the development and popularity of light detection and ranging (LiDAR) brings new energy to the tracking research community, whose research subjects have expanded from classical visual tracking \cite{Kristan2015TheVO, Bertinetto2016FullyConvolutionalSN, Kristan2016ANP} to challenging 3D object tracking in point clouds. In general, 3D tracking tasks can be roughly divided into 3D single object tracking (3D SOT) and 3D multiple object tracking (3D MOT). The former is designed to locate the object arbitrarily specified in the first frame, while the latter concentrates on finding all class-specific objects through the detection-and-association mechanism. In following discussion, the research objective will focus on solving some pain points existing in popular 3D single object tracking paradigms.

% The essential discrepancy between them is that 3D SOT requires a fixed number of targets (\ie, one) but any categories, whereas 3D MOT usually deal with the scenarios of any number of targets but fixed categories.
\begin{figure}[t]
  \begin{center}
     \includegraphics[width=1\linewidth]{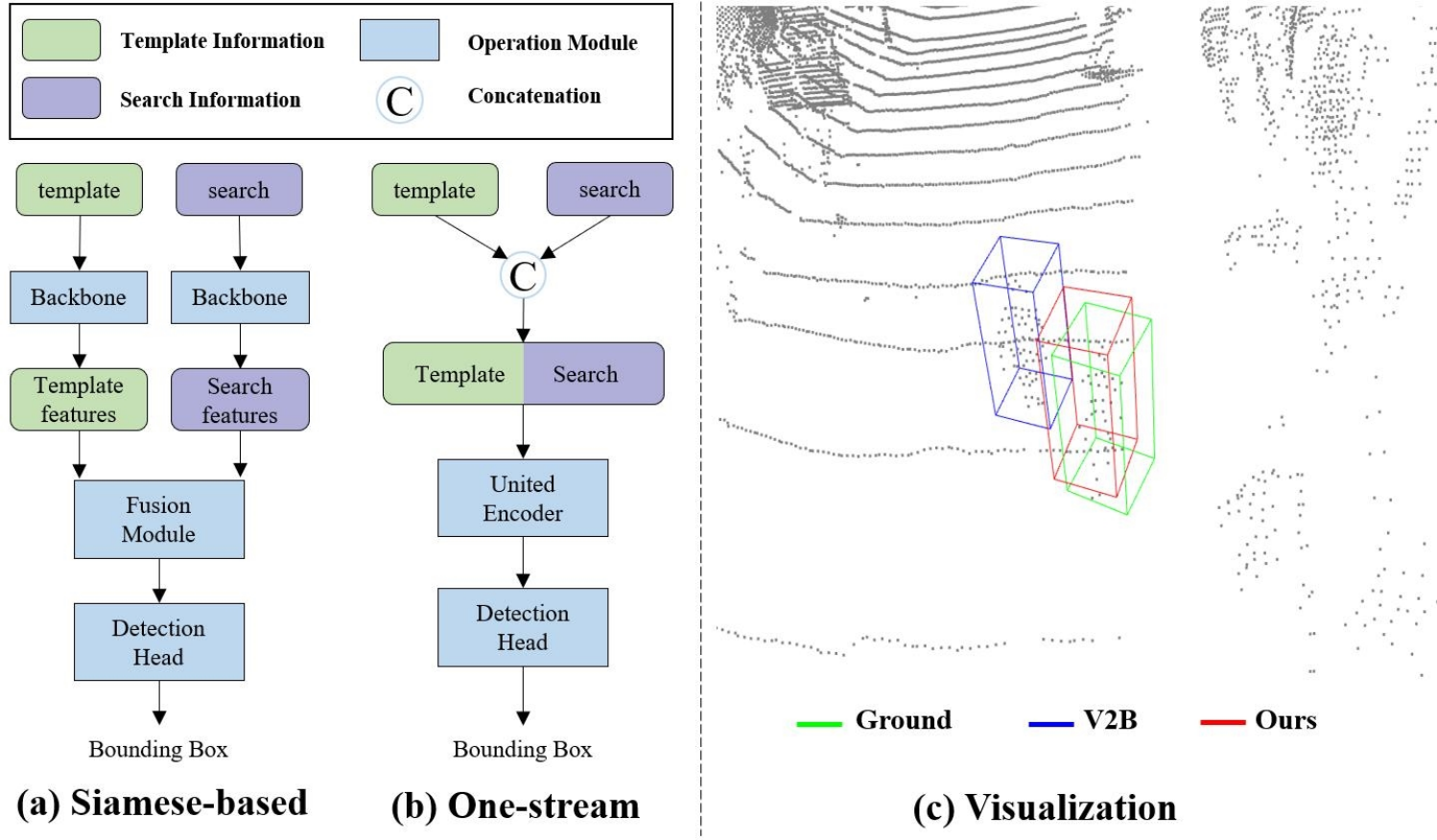}
  \end{center}
  \setlength{\abovecaptionskip}{-0.15cm}
  \setlength{\belowcaptionskip}{-0.cm}
  \caption{Comparison between Siamese-based and one-stream frameworks attached their visualization results in case of interference from other similar targets. (a) The basic process of classical Siamese methods, \emph{e.g.} P2B, V2B. (b) The novel approach with one-stream frameworks. (c) The visualization result of V2B and our method, where ours captures the target pedestrian while V2B is misled by the interferes.}
  \label{fig:introduce}
\end{figure}

%3D SOT aims to estimate the state of the target (\ie center, size, and yaw angle) in other frames with an arbitrary given object tightly enclosed by a bounding box in the first frame. 
There are three mainstream schemes for solving 3D SOT problems in this communities. The first one deals with this task using purely appearance model, such as SC3D~\cite{Giancola2019LeveragingSC}, which learns a general feature matching as a 2D counterpart of SiamFC~\cite{Bertinetto2016FullyConvolutionalSN}. Nevertheless, the model cannot conduct end-to-end train limited by multi-frames polymerization and feature matching. The second is the family of motion-based methods, like MM-track~\cite{Zheng2022Beyond3S}, SETD~\cite{Tian2021LearningTI}, Register-driven~\cite{Jiang2022PointCR}, which utilized the aligned features or refine the coarse bounding box with target relative motion. Although the adoption of motion clues has greatly improved tracking accuracy, the correct prediction of target relative motion in the sparse scenes is also an arduous task. 
The last one based on feature fusion, relying on its lean structure and easiness to train, dominates the trend of this task. These works integrate the 3D detection head (VoteNet~\cite{Qi2019DeepHV}, VoxelNet \cite{Zhou2018VoxelNetEL}) hinged on a powerful feature embedding between the template and search area, typical examples include P2B \cite{Qi2020P2BPN}, BAT \cite{Zheng2021BoxAwareFE}, PTT \cite{Shan2021PTTPM}, V2B \cite{Hui20213DSV} and so on. Even though leaving an impressive performance, all above frameworks still struggle with two dilemmas.

On the one hand, Siamese-based methods suffer from two inherent barriers, which are caused by independent feature extraction of the template and search region. Firstly, the features extracted by Siamese backbone are only able to make category-level rather than instance-level distinction due to the absence of the template information, resulting in insufficient discernment between targets and background distractors. For example, when several similar objects simultaneously appear in the search region, Siamese-based trackers may be misled by the interferes, as shown in Figure.~\ref{fig:introduce}. Secondly, Siamese-based frameworks always struggle between efficiency and effect. In fact, the performances of Siamese-based methods are vulnerable to the feature fusion between the template and search region. The current fusion modules have evolved from the single MLP augmentation \cite{Qi2020P2BPN} to the complicated and recurrent Transformer \cite{Hui20223DST}. Undoubtedly, a powerful feature fusion module is a guarantee of the ultimate perfect tracking, but also means higher computational costs. Therefore, it is non-trivial for Siamese-based methods to both refine design for accuracy and compact model for speed. To tackle these two problems, we propose a radically novel and efficient one-stream framework for 3D SOT, the structure is shown in Figure.~\ref{fig:2_Network}. Benefiting from one-stream network as a bridge between the template and search region, the target information can flow to search region such that features with more discrimination can be acquired under the guidance of the template. In addition, owing to abandonment of feature fusion module, the proposed tracker converge faster and no need to tangle in efficiency.

On the other hand, the vast majority of existing methods forward under the class-specific setting, which overlook the inherent merit of 3D SOT. As described above, the essential discrepancy between them is that 3D SOT requires a fixed number of targets (\ie, one) but any categories, whereas 3D MOT usually tackles the scenarios of any number of targets but fixed categories. In other words, single object tracker should be able to find target with the given template no matter whether the category was known or not. However, current methods rely heavily on training on homologous samples. For instance, it is inevitably to train the model on car category if a reliable car tracker is expected, which is obviously against the principle of 3D single object tracking. Taking this issue into account in our work, we conduct experiments under category-agnostic settings which can intuitively reflect the generalization ability of the evaluated trackers and the extensive experimental results on prevalent datasets convincingly illustrate that the proposed method can trace more accurately to instance than previous methods.

In summary, our main contributions are as follows:
\begin{itemize}
  \item A radically novel one-stream tracking framework base on Transformer is proposed, which can embed template information into search region during features extraction such that more discriminating attributes can be acquired.
  \item We introduce a multi-scale feature extraction module to ensure both local geometrical information and global semantic information complementary to each other,  
  which is characterized by reverse directional feature propagation that aggregates information from shallow to deep layers.
  \item We conduct some representative experiment on KITTI \cite{Geiger2012AreWR} and nuScence \cite{Caesar2020nuScenesAM} datasets. The proposed method considerably reduces the computational cost and thus achieves high training and inference speed, without the loss of the tracking accuracy under the class-specific setting compared to the state-of-the-art methods. And more notably, our one-stream design for 3D tracking obtains better performance under the class-agnostic setting.
\end{itemize}

%% file: related_work.tex
\section{Related Work}

\begin{figure*}[ht]
  \centering
  \begin{overpic}[width=2.0\columnwidth,tics=10]{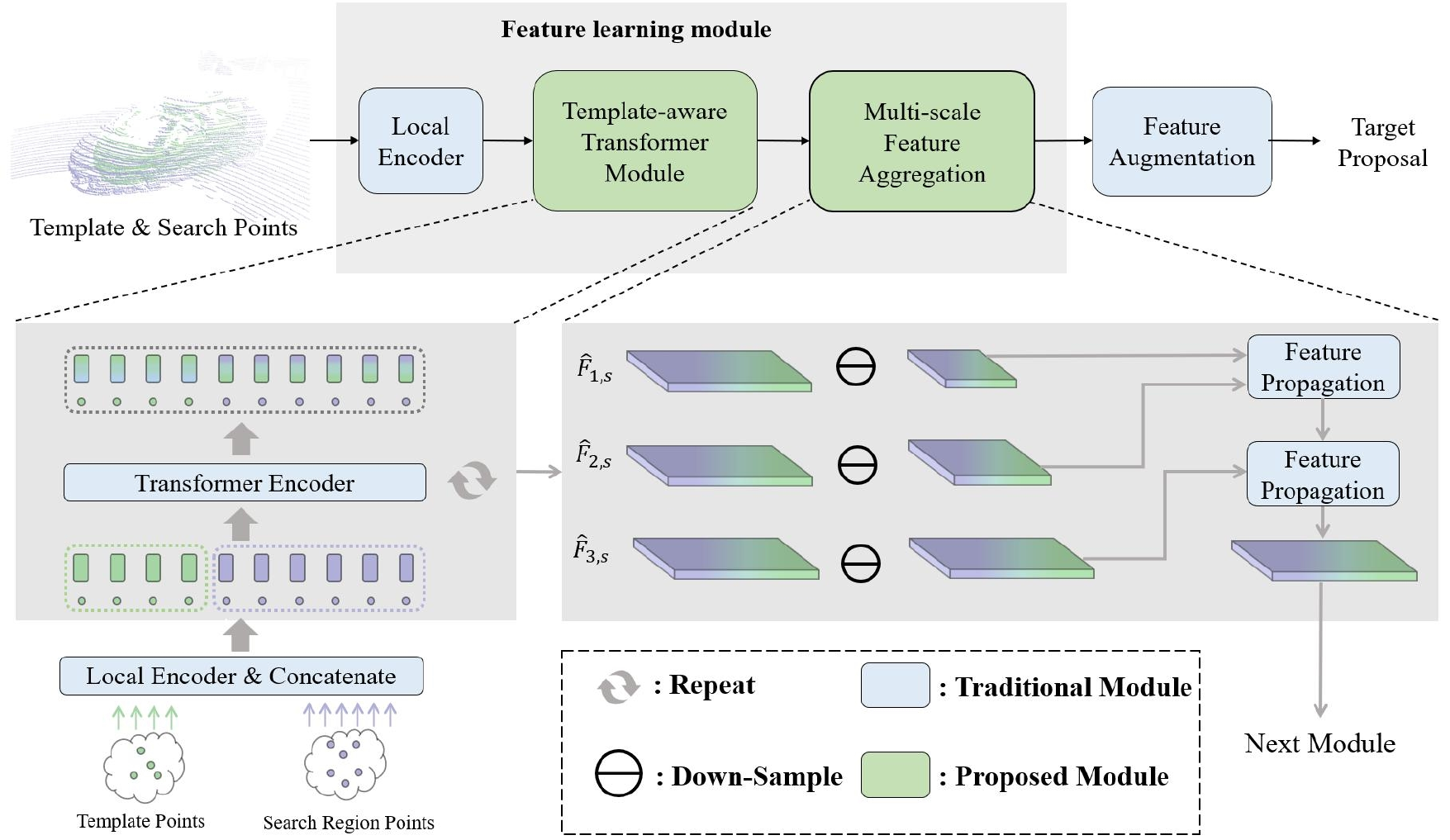}  
  \end{overpic}
  \caption{The overview of the proposed OST. After concatenating the template and the search region, we send it into the one-stream network consisting of local encoding module, template-aware Transformer module, and multi-scale feature aggregation module. In this way, we can generate template-aware features of search region. Afterwards, following the feature augmentation by the segmentation prior, we voxelized feature tensor for proposal generation. The overall process is at the top of this figure, and the module details are at the bottom.}
  \label{fig:2_Network}
  \vspace{-7pt}
\end{figure*}

\subsection{3D Siamese Tracking}
 Shape Completion 3D (SC3D) network \cite{Giancola2019LeveragingSC} is the pioneering work in this category, in which Siamese tracker encodes the targets and search regions into latent representations, then the target can be determined by comparing similarity of latent representations among hand-crafted proposals. After SC3D, Point-to-Box (P2B) network \cite{Qi2020P2BPN} is another milepost. It uses a Siamese PointNet++ \cite{Qi2017PointNetDH} to encode the template and search region, then merges the template information into the down-sampled search region by target-specific feature augmentation module, and eventually a detection head based on VoteNet \cite{Qi2019DeepHV} consumes such augmented search region to regress the most appropriate bounding box. P2B solves the problems that SC3D cannot perform end-to-end training and candidate generation. Subsequently, in light of its potential improvement room, researchers proposed some variants, such as BAT \cite{Zheng2021BoxAwareFE} which embeds information of the bounding box into search seeds during feature fusion process and MLVSNet \cite{Wang2021MLVSNetMV} which conducts multi-level voting to deal with sparsity in point clouds. In spite of outstanding performance at that time, its results are not state-of-the-art compared with the current cutting-edge methods. The latest Siamese tracking methods replace voting head by Bird's-Eye-View (BEV) head \cite{Luo2022ExploringPF, Oleksiienko2022VPITRE, Zarzar2019EfficientBE, Cui20213DOT, Liu2022BEVFusionMM, Hui20213DSV, Pan20213DOD}, in which points are projected into structured 2D feature maps and the maps will be used to regress bounding boxes. V2B \cite{Hui20213DSV} transfers fusion features from point to BEV via voxelization and max-pooling, then BEV features will be sent into some 2D-Convolution layers to regress target bounding box. LTTR \cite{Pan20213DOD} transfers the template and search region features into BEV maps separately, and then conducts feature fusion on BEV through sheer force of Transformer. As discussed above, all mentioned approaches perform well under the class-specific setting but seldom unattended under category-agnostic settings \cite{Tian2022TowardsCT}. To gap this challenge, we propose a one-stream framework for 3D single object tracking which can capture target-wise rather than category-wise attributes so that whose accuracy far exceeds other baseline methods under category-agnostic settings.

\subsection{Transformer in Point Cloud}
Benefiting from Transformer's powerful encoding capacity, the precision of 2D visual tasks have been achieved great improvement \cite{Carion2020EndtoEndOD, Dosovitskiy2021AnII, Liu2021SwinTH}. Nonetheless, Transformer was not widely used in point cloud tasks over a long previous period of time owing to irregularity of 3D point cloud data. With the deepening of research, more and more works in point cloud tasks based on Transformer are emerging in the near future \cite{Cui20213DOT, Pan20213DOD, Hui20223DST, Zhang2022CATDetCA, Shan2021PTTPM}. Pan \etal \cite{Pan20213DOD} proposed Pointformer method, which consists of a local Transformer to learn context-dependent region features, a global Transformer to learn context-aware representations, and a local-global Transformer to integrate local features with global features from higher resolution. Hui \etal \cite{Hui20223DST} present a 3D Siamese Transformer network (STNet) for robust cross correlation learning between the template and the search area of point clouds. It learns multi-scale features to guide 3D tracking and produces remarkable results in large-scale outdoor datasets. Shan \etal \cite{Shan2021PTTPM} proposed Point-Track-Transformer (PTT) consisting of three blocks for feature embedding, position encoding, and self-attention feature computation and embedded it into P2B framework whose performance substantially surpasses fundamental P2B. In this work, we illustrate a novel multi-scale feature aggregation module following Transformer framework and prove its effectiveness in KITTI \cite{Geiger2012AreWR}, nuScenes \cite{Caesar2020nuScenesAM} datasets.

\subsection{2D One-Stream Tracking}
In recent years, some 2D one-stream frameworks \cite{Ye2022JointFL, Xie2022CorrelationAwareDT} have been proposed to perform their unique character in 2D object tracking. Ye \etal \cite{Ye2022JointFL} released some Siamese-based framework's problem: the extracted features lack the awareness of the target and have limited target-background discriminability. To tackle these issues, Ye \etal introduced a one-stream tracking framework which can bridge the template-search image pairs with bidirectional information flows. Single Branch Transformer (SBT) \cite{Xie2022CorrelationAwareDT} suppresses non-target features and obtain instance-varying features by extensively matching the features of the two images through one-stream backbone. To sum up, one-stream frameworks possess the ability to extract more discriminative features than ordinary Siamese frameworks. Inspired by these works, we migrate this attribute to 3D single object tracking with elaborately designed template-aware Transformer and multi-scale feature aggregation for point clouds.

%% file: proposed_method.tex
\section{Method}
In 3D single object tracking, a tracker aims to locate the target at each frame of point cloud sequence. 
In general, the target is specified in the first frame to guide the tracking procedure in the following frames. 
Mathematically, given the template $P_t$ and search region $P_s$, which are composed of $N_t$ and $N_s$ coordinates $(x, y, z)$ respectively, a tracker is supposed to accurately predict the state of the specified object in the search region $P_s$. 
We make use of a 3D bounding box to represent the state of the target, which is parameterized by center $(x,y,z)$, size $(l,w,h)$, and yaw angle $\theta$. Considering the outdoor scenes captured by LiDAR generally dose not deform the physical size of the rigid target, the tracking problem can be formulated as the following,   
\begin{equation}\label{tracker}
\begin{aligned}
   \{x,y,z,\theta \} = \Phi(P_t, P_s),
\end{aligned}
\end{equation}
where $\Phi$ represents the tracking model. 
%介绍两点，第一，siamese的问题：效率和准确率平衡，第二，忽略了类无关：目标感知的特征缺乏。提出了,,,,方法，能够通过所提方法的什么的特点解决第一点，又能通过所提方法什么特性解决第二点。
Many existing algorithms have been proposed to toward for this problem. The Siamese-network based method focus on learning of shape features and applying correlation operation between the template and search region. This needs a high time consumption in despite of achieving impressive performance. 
Besides, the current methods focus only on tracking those object categories seen during training. We deem that they lack the ability to learn discriminative features to capture target information that is arbitrarily specified in the first frame. To overcome this shortcoming, we propose a novel method with the philosophy of one-stream. The proposed method can not only reduce computing costs by integrating feature extraction with relation modeling, but also enhance the ability to learn discriminative features by building an information flow between the template and search region.

% The proposed method includes three core parts: Jointing-based transformer module (JTM), multi-scale feature aggregation (MFA), as well as a specific loss. An overview of the model is shown in Fig. \ref{fig:2_Network}.  In the subsection, we first introduce our one-stream architecture which is composed of cascaded JTM, and then present the details of MFA, finally the task-specific loss is described. 

% 调换一下顺序？让上下文 通顺
An overview of the model is shown in Figure~\ref{fig:2_Network}. It includes three core parts: template-aware Transformer module (TTM), multi-scale feature aggregation (MFA), as well as a specific loss. In the subsection, we first introduce our one-stream architecture which is composed of cascaded TTM, and then present the details of MFA, finally the task-specific loss is described.

\subsection{One-stream Architecture}

Most current methods are based on Pointnet++\cite{Qi2020P2BPN, Zheng2021BoxAwareFE, zhou2022pttr} or 3D sparse convolution~\cite{Cui20213DOT} to extract point cloud features, and continue the practice of Siamese networks with shared parameters as 2D tracking did. The core to this framework must rely on a powerful feature fusion module to embed the template information into the search area, allowing the generation of features that discriminate strongly in the foreground and background. Such fused features will further facilitate the subsequent verification of the target by the detection head. Different from these conventional workflow, we introduce a one-stream architecture, the standpoint behind which lies in two aspects: One is that the correlation between the template and search region is discarded for high efficiency; the other one is the jointly feature learning for the $P_t$ and $P_s$ using the Transformer instead of PointNet++ and sparse convolution. 

% 下面这段话是说的one-stream吗？感觉像是在为多尺度铺垫？
% Concretely, for the basic module of the transformer, we take both semantic information and spatial relationship , in which the former determines whether the generated proposals match well with the template and the latter plays a critical role in the motion regression of the target. That is, with the deepening of the Transformer, the semantic of the feature map becomes more stronger, whereas the spatial information is reduced inevitably. Bear this issue in mind, we design a specific multi-scale feature aggregation method to supplement the spatial information after Transformer learning.

\begin{figure}[t]
  \begin{center}
     \includegraphics[width=0.9\linewidth]{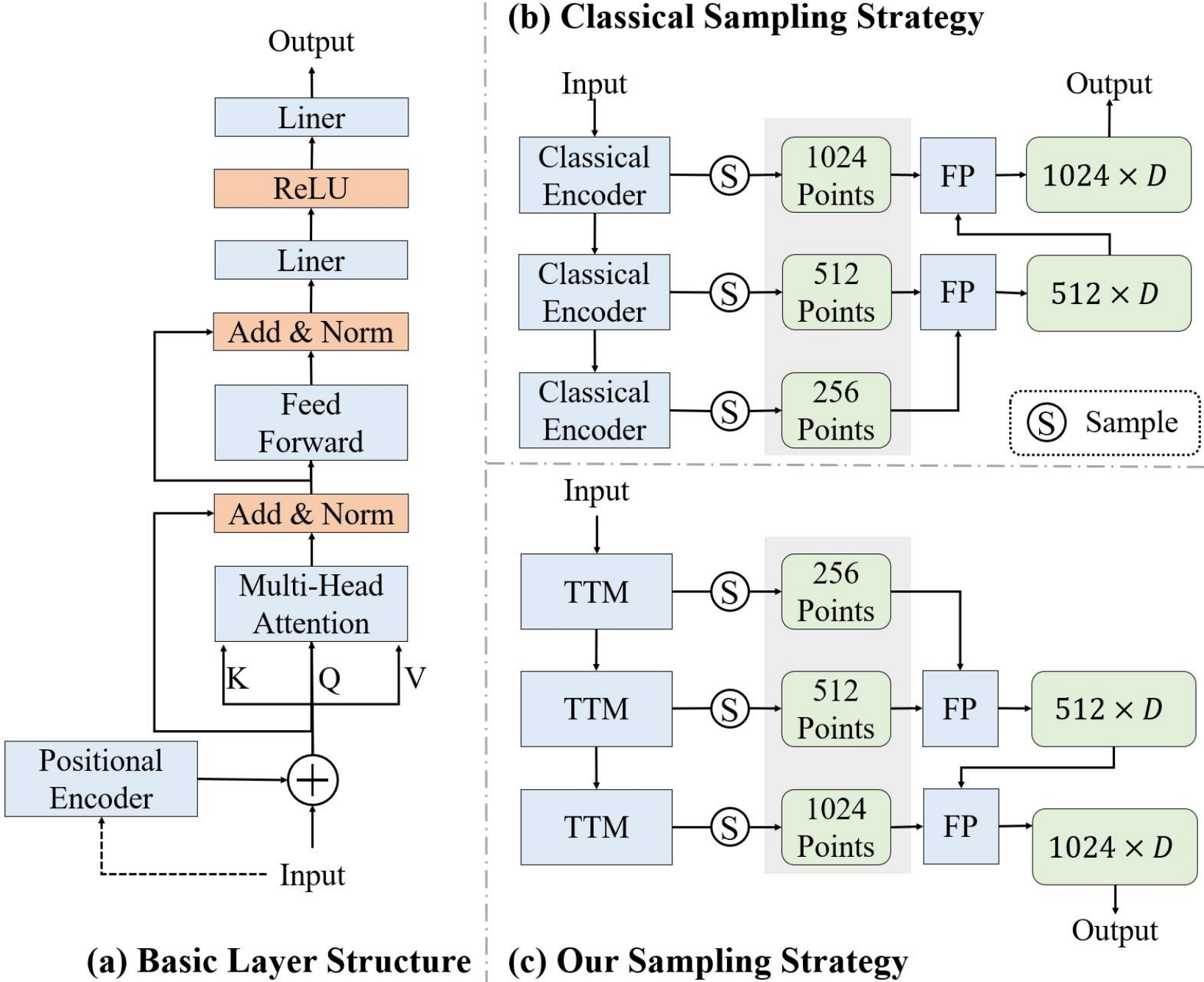}
  \end{center}
  \setlength{\abovecaptionskip}{-0.15cm}
  \setlength{\belowcaptionskip}{-0.cm}
  \caption{The architecture of our Transformer encoder and two distinctive aggregation strategies. (a) The basic process of Transformer in our method. (b) The common steps for multi-scale aggregation. (c) Inverse sampling strategy for multi-scale aggregation with our template-aware Transformer module (TTM).}
  \label{fig:transformer&multi-scale}
\end{figure}

\textbf{Template-aware Transformer Module.} 
Our one-stream network is composed of template-aware Transformer encoder module. Specifically, we sample 512 points $P_{t} \in \mathbb{R}^{512 \times 3}$ in the template  and 1024 points $P_{s} \in \mathbb{R}^{1024 \times 3}$ in the search region, and we feed the points in template \begin{math} P_{t} \end{math} and search area $P_{s}$ to a one-stream feature extraction backbone. First, since the raw coordinate of each point cannot describe its local structure, we thus adopt graph convolution neural network (GCN) \cite{welling2016semi} to encode spatial features in the corresponding neighbor. And the resulting features of the template and search region are represented as $F_{t} \in \mathbb{R}^{512 \times D} , F_{s} \in \mathbb{R}^{1024 \times D}$, respectively, in which $D$ indicates the feature dimension. Next, as the input of the one-stream network, we need to concatenate both coordinates and features of the template and search region, yielding $P_c = [P_t; P_s]$ and $F_c = [F_t; F_s]$. Then, the Transformer encoder $\phi$ takes the $P_c$ and $F_c$ as inputs to establish a interactive flow between $F_t$ and $F_s$. We can simply formulate this procedure as
\begin{equation}\label{eq:transformer}
    \hat{F}_c=\phi(P_c,F_c),
\end{equation}
where $\hat{F}_c$ is a template-aware feature produced by one Transformer block. 

% Because the sort of the point-wise feature in the tensor doesn't change during the transformer, the output also can be written as

% \begin{small}
% \begin{equation}\label{tau_2}
% \begin{aligned}
%     & \relax[\hat{F}_{t};\hat{F}_{s}]=\phi([P_{t}; P_{s}],[F_{t};F_{s}])
% \end{aligned}
% \end{equation}
% \end{small}

The inside mechanism of the Transformer encoder $\phi$ is illustrated in the Figure~\ref{fig:transformer&multi-scale}-(a). In our design, we iteratively implement this block with $\tau$ times, mathematically, it can be written as:
\begin{equation}
 \hat{F}_c^{(i)} = \phi^{(i)}(P_c^{(i-1)},  \hat{F}_c^{(i-1)}),  
\end{equation}
where the superscript $i \in \{1, 2, \cdots, \tau\}$ represents the $i$-th Transformer block and $\hat{F}_c^{(0)} = F_c$. In our method, we perform it iteratively three times, \ie,$\tau=3$. Note that the output in the previous iteration will replace the input in the next iteration. In the ablation study, we will fully explore the impact of the number of  $\phi$. 

%介绍这个transformer。 ??
% Why transformer can create a flow of information between t and s
%上节我们说到了，transformer 可以在模板和搜索区域之间建立信息流，自注意力机制在其中起到了至关重要的作用.
\textbf{Analyze of Transformer module.} As we mentioned in the previous section, the Transformer in our method can establish information flow between the template and the search region. In fact, the self-attention mechanism in Transformer plays an important role. For simplification, we take the equation (\ref{eq:transformer}) as example and further analyze the intrinsic reasons for it. The output of self-attention mechanism in our method can be written as:
\begin{equation}
   \begin{aligned}
    \hat{F}_c &=\phi(P_c, F_c) \\ 
        &= Softmax(\frac{Q_c K_c^{T}} {\sqrt{d}}) \cdot V_c,\\
    \end{aligned}
\end{equation}
where $d$ is the feature dimension of each attention head; $Q_c$, $K_c$, and $V_c$ are query, key and value matrices Transformed from $F_c + PE(P_c)$ by the linear layers. Herein, $PE(\cdot)$ is the a learnable position embedding. 

Since the $P_c$ and $F_c$ derive from the concatenation of the template and search region, we can also decompose the  $Q_c$, $K_c$, and $V_c$ as the following formulation:
\begin{equation}
\begin{aligned}
    Q_c &= [Q_t; Q_s], \\
    K_c &= [K_t; K_s], \\
    V_c &= [V_t; V_s],
\end{aligned}
\end{equation}
where the subscripts $t$ and $s$ denote the template and search region separately. 

Substituting the $Q_c$, $K_c$, and $V_c$ with the above equation, we can further obtain
\begin{small}
\begin{equation}
\begin{aligned}
        \hat{F}_c&= Softmax(\frac{[Q_{t};Q_{s}][K_{t};K_{s}]^{T}} {\sqrt{d_{k}}})\cdot [V_{t};V_{s}]\\
        &\triangleq [W_{tt}, W_{ts}; W_{st}, W_{ss}] \cdot [V_{t};V_{s}] \\
        &= [W_{tt}V_{t} + W_{ts}V_{s}; W_{st}V_{t} + W_{ss}V_{s}], \\
\end{aligned}
\end{equation}
\end{small} 
where $[W_{tt}, W_{ts}; W_{st}, W_{ss}]$ is the attention weight, and the $W_{ab}, a,b \in \{t,s\}$ is a measure of similarity between $a$ and $b$. 
It can be seen that this one-stream structure based on the Transformer has essentially fulfilled the extraction of the template (\ie, $W_{tt}V_{t}$) and the search region (\ie, $W_{ss}V_{s}$), and realized the information interaction between them (\ie, $ W_{ts}V_{s}$ and $ W_{st}V_{t} $). Therefore, we can leverage this template-aware features to complete the verification of the target, without extra correlation module.  
% \begin{small}
% \begin{equation}\label{tau_3}
% \begin{aligned}
%     \hat{F}_{s} = W_{st}V_{t} + W_{ss}V_{s},
% \end{aligned}
% \end{equation}
% \end{small}

% As results, $W_{st}V_{t}$ establish the relationship between template and search region based on similarity $W_{st}$, and $W_{ss}V_{s}$ aggregate the feature inside the search region based on similarity $W_{ss}$. 

\subsection{Multi-scale Feature Aggregation} 
In order to make spatial and semantic information complement each other for tracking task, we elaborately design a multi-scale feature aggregation strategy. In fact, the semantic information determines whether the generated proposals match well with the template, while the spatial relationship plays a critical role in the motion regression of the target. With the deepening of the Transformer, the semantic of the feature map becomes more stronger, whereas the spatial information is reduced inevitably. Bear this issue in mind, we design a specific multi-scale feature aggregation method to supplement the spatial information after Transformer learning.

For the input point cloud pairs (template and search region), their coordinates are $P_t \in \mathbb{R}^{512 \times 3}$, $P_s\in \mathbb{R}^{1024 \times 3}$, and their features are $F_{t}$, $F_{s}$. After the template-aware Transformer module, we can calculate the feature set $[\hat{F}_{c}^{(1)},\hat{F}_{c}^{(2)},\hat{F}_{c}^{(3)}]$ from different layers of Transformer. According to Equation (5), by truncating the $\hat{F}_{c}^{(i)}$, we can obtain a template-aware feature $\hat{F}_s^{(i)}$ for the search region in each Transformer block. Next, we down-sample search region points from each Transformer layer by farthest point sampling method. Different from the common method, We sample $256$ points $P_{s}^{(1)}$ in the first TTM layer and $512$ points $P_{s}^{(2)}$ in the second Transformer layer. Then we gather the feature $\hat{ \mathcal{F} }_s^{(1)}$, $\hat{\mathcal{F}}_s^{(2)}$ of sampled points from $\hat{F}_{s}^{(1)}$ and $\hat{F}_{s}^{(2)}$. As shown in Figure~\ref{fig:cost}, we aggregate different scales features of search region using Feature Propagation \cite{Qi2017PointNetDH} and obtain the final feature $\widetilde{F}^{(3)}_{s}$ from this module.
\begin{equation}\label{tau_7}
\begin{aligned}
    &\widetilde{F}^{(2)}_{s} = FP(\hat{\mathcal{F}}_s^{(1)},\hat{ \mathcal{F} }_s^{(2)}), \\
    &\widetilde{F}^{(3)}_{s} = FP(\widetilde{F}^{(2)}_{s},\hat{F}_s^{(3)}). 
\end{aligned}
\end{equation}

Multi-scale features in different layers capture the information at different scales of the point cloud. In our method, as the features flowing in the template-aware Transformer module, the semantic information is stronger and the spatial information weaker. Our MFA module embeds the spacial information of the front two layers into the final layer. Through multi-scale feature aggregation, spatial information and semantic information can be effectively balanced.

\subsection{Loss Function}
%Then, We voxelize the features ${F}_{0}^{s},{F}_{1}^{s},{F}_{2}^{s}$ from different layers separately, followed by concatenating the voxelized feature tensors with size $[B,D_0+D_1+D_2,L,W,H]$ . 
% 我们参考 V2B 中的Voxel-to-bev 方法作为我们的region proposal network,首先将搜索区域内的点体素化, where the feature map with size of [B,D,L,W,H] ,将体素映射到鸟瞰图上，where the feature map with size of [B,D,W,H].然后在bev上进行提案预测。

\begin{figure}[t]
  \begin{center}
     \includegraphics[width=1\linewidth]{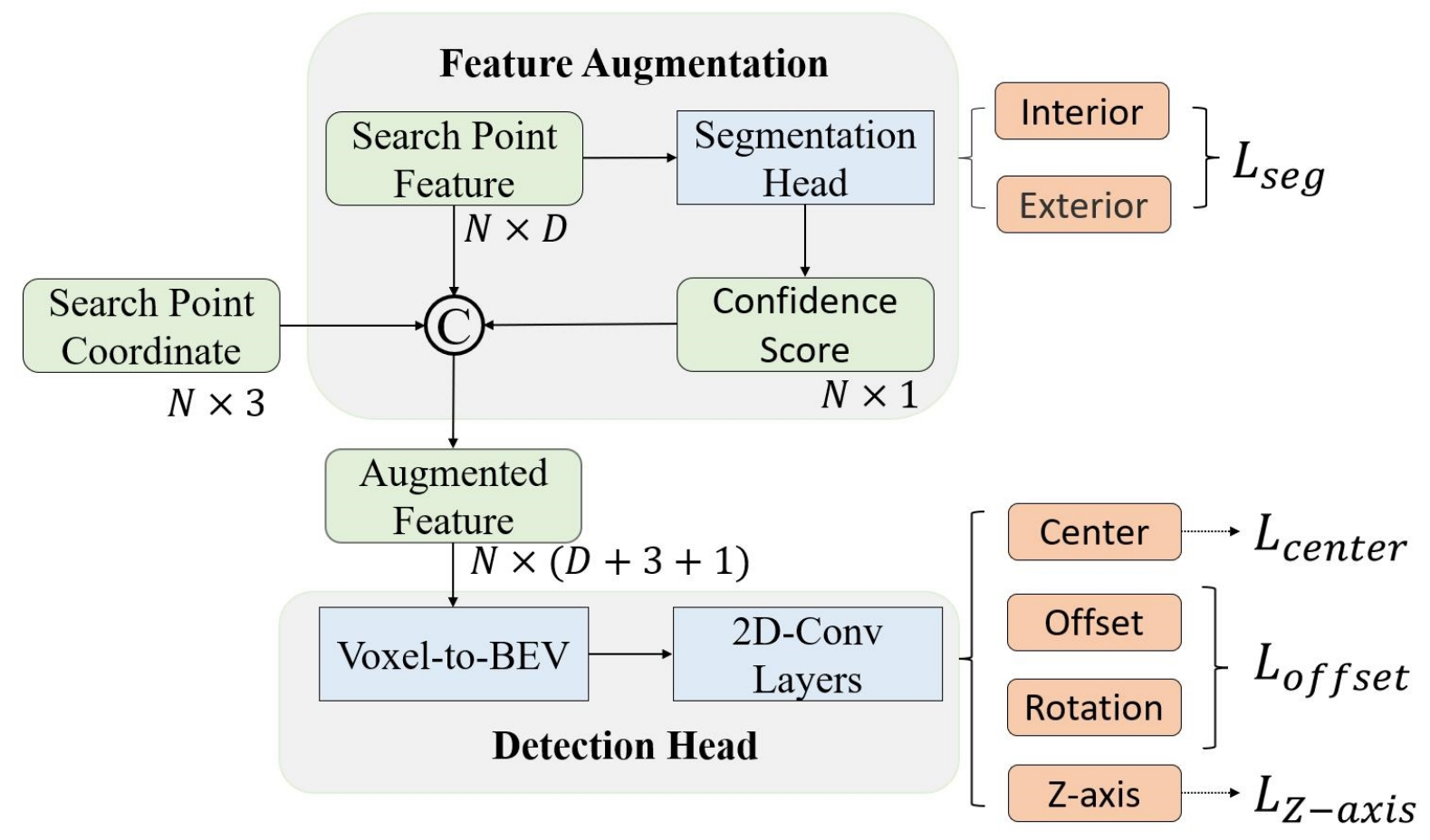}
  \end{center}
  \setlength{\abovecaptionskip}{-0.15cm}
  \setlength{\belowcaptionskip}{-0.cm}
  \caption{The architecture of feature augmentation and detection head. The feature augmentation is guided by the segmentation head which provides interior and exterior cues. The detection head is used to regress the bounding box following V2B.}
  \label{fig:seg&head}
\end{figure}

We take the Voxel-to-BEV method in \cite{Hui20213DSV} as our region proposal network. It first voxelizes the points in the search region with size of $[L,W,H]$. Remaining the original feature dimension, we can get the feature map with size of $[D,L,W,H]$. Then these voxels are projected onto the bird's eye view (BEV), generating the feature map with size of $[D,W,H]$. In the end, proposal prediction is performed on the BEV feature map. More importantly, to improve the performance of this subnetwork, we introduce a new task-specific loss which considers the segmentation regularization to promote the verification of the target. The insight behind this design is that the points lying on the surface of the target is a critical clue which can provide the discriminativeness for the subsequent target location. Therefore, we enhance semantic features of the search region using segmentation instead of completion. In the Section~\ref{ablation}, we will investigate this effect comprehensively.

%其中预测包括bev上中心位置离散化回归，离散中心到连续中心的offset回归，z轴中心位置的回归。
The final prediction head includes the center loss $L_{center}$ which controls the regression of the center position on the BEV, the offset loss $L_{offset}$ which constrains the movement vector of the target, the Z-axis loss $L_{z-axis}$ which reflects the difference in height of the target, as well as the segmentation loss $L_{seg}$ which promotes the target verification by the discriminative score. 
The overall loss function can be written as:
\begin{equation}\label{tau_11}
\begin{aligned}
L_{total}=  \lambda_1 L_{seg} + \lambda_2 L_{center} + \lambda_3 L_{offset} + \lambda_4 L_{z-axis},
\end{aligned}
\end{equation}
where $\lambda_1$, $\lambda_2$,  $\lambda_3$ , and $\lambda_4$ are the hyper-parameter for segmentation, center regression, movement vector regression, and z-axis height regression, respectively. In our experiments, we empirically set these parameters as $\lambda_1=1.0$, $\lambda_2=1.0$,  $\lambda_3=1.0$ , and $\lambda_4=2.0$. In the following, we will describe each loss term in detail. 

\begin{table*}[htb]
  \caption{Performance comparison with advanced methods. The Success/Precision are reported and the best results are highlighted in bold.}
  \label{tab:class_special_result}
  \centering
  \setlength{\tabcolsep}{1.5mm}{
  \begin{tabular}{lccccc|ccccc}
    \toprule
    \multirow{3}{*}{Methods} & & & \multicolumn{1}{c}{KITTI}  & & & & & \multicolumn{1}{c}{nuScenes} &  \\
    \cmidrule(lr){2-6} \cmidrule(lr){7-11}
              & Car         & Pedestrian   & Van         & Cyclist & Mean       & Car     & Pedestrian   & Truck  & Bicycle & Mean\\
              & 6424       &6088      &1248       &308      &14068    & 15578      &8019       &3710    &501    &27808\\
    \midrule
    SC3D    & 41.3/57.9 & 18.2/37.8 & 40.4/47.0 &  41.5/70.4 & 31.2/48.5 &25.0/27.1 & 14.2/16.2 &\textbf{25.7/21.9} &17.0/18.2 &21.8/23.1   \\
    P2B & 56.2/72.8 & 28.7/49.6 & 40.8/48.4 & 32.1/44.7 &42.4/60.0  & 27.0/29.2 & 15.9/22.0 & 21.5/16.2 & 20.0/26.4 &22.9/25.3 \\
    MLVSNet &56.0/74.0 & 34.1/61.1 &52.0/61.4 &34.3/44.5 & 45.7/66.6 & -- & -- &-- &-- & -- \\
    %DSDM &  &  &  &  &  &\\
    %SETD &  &\\
    LTTR & 65.0/77.1 &33.2/56.8 &35.8/45.6 & 66.2/89.9 &48.7/65.8 & -- &-- &-- & -- &-- \\
    BAT &60.5/77.7 &42.1/70.1 &52.4/67.0 &33.7/45.4& 51.2/72.8 & 22.5/54.1 &17.3/24.5 &19.3/15.8 &17.0/18.8& 20.5/23.0 \\
    PTTR & 65.2/77.4 & 50.9/81.6 &52.5/61.8 &65.1/90.5 &58.4/77.8 & -- &-- &-- &-- &--\\
    PTT & 67.8/81.8 & 44.9/72.0 &43.6/52.5 &37.2/47.3 &55.1/74.2 & -- & -- &-- &-- &-- \\
    V2B & 70.5/81.3 &48.3/73.5 & 50.1/58.0 &40.8/49.7 &58.4/75.2  & 31.3/35.1 & 17.3/23.4 & 21.7/16.7 &\textbf{22.2}/19.1 &25.8/29.0 \\
    STNet &\textbf{72.1}/84.0 &49.9/77.2 &48.0/\textbf{70.6} &\textbf{73.5/93.7} &\textbf{61.3}/80.1  & \textbf{32.2/36.1} & \textbf{19.1/27.2} &22.3/16.8 &21.2/\textbf{29.2} & \textbf{26.9/30.8} \\
    \textbf{Ours} & 72.0/\textbf{84.2} &\textbf{51.4/82.6}	& \textbf{57.5}/68.2 & 49.2/60.4 & \textbf{61.3/81.6} & 26.6/28.0 & 14.8/16.2 &19.3/14.7	&  17.0/16.4& 22.1/22.6 \\
    \bottomrule
  \end{tabular}}
\end{table*}

\textbf{Center regression on BEV.} Following V2B~\cite{Hui20213DSV}, the target is located by the peak of the heatmap $H \in R^{ H \times W \times 1}$. we project the 3D ground-truth from the bird's eye view, and can naturally generate the 2D target center and rectangle for the regression of detection head. For pixel $p_{i,j}$ in BEV, (1). if the pixel is the 2D ground truth center,  $H_{i,j}=1$. (2). if the Pixel is not the 2D ground truth center but within the 2D ground truth bounding box , $H_{i,j} = \frac{1}{1+d} $ where $d$ is the distance from this pixel to 2D ground truth center. (3). if the pixel is out of the 2D ground truth bounding box, $H_{i,j}=0$. The loss can be calculated by modified focal loss~\cite{lin2017focal, law2018cornernet, ge2020afdet} as follows:
\begin{equation}\label{tau_8}
\begin{aligned}
L_{center}= & - \sum \{I[H_{ij}=1] \cdot (1-\hat{H}_{ij})^\alpha log(\hat{H}_{ij}) \\
             & + I[H_{ij} \ne 1] \cdot (1-H_{ij})^{ \beta } (\hat{H}_{ij})^\alpha log(1- \hat{H}_{ij})\},
\end{aligned}
\end{equation}
where $\hat{H}_{i,j}$ is the score whether the pixel is the center point predicted by bev head.  Where $ I(cond.) $ is the indicator function. If $cond.$ is true, then $I(cond.) = 1$, otherwise $0$. Besides, we set $\alpha = 2 $ and $\beta = 4 $ in all experiments.

\textbf{Offset regression.}
%$L_{center}$ 将目标中心约束到二位离散中心，对于精确预测物体中心是不准确的。在回归时，我们考虑回归到连续的2D地面真实中心，即回归到三维物体中心映射到BEV上的精确位置而非像素点。We train the offset using L1 loss[] in the time train the rotation
$L_{center}$ constrains the target center to a discrete center on BEV, which is inaccurate for predicting the target center. 
We consider regressing the target center from discrete to the continuous 2D ground truth center. We regress the offset from pixel center position to the continuous position that the 3D object center mapped to the BEV. % 连续坐标 非 离散坐标。
We select a square area with the radius $r$ around object center pixel in the offset regression map. And we train the offset using $L_{1}$ loss~\cite{ren2015faster} meanwhile train the rotation.
\begin{small}
\begin{equation}\label{tau_9}
\begin{aligned}
L_{offset}= \sum^{r}_{ \delta = -r} \sum^{r}_{\gamma = -r} \left | \hat{O}_{\widetilde{c}+(\delta + \gamma)} -[c-\widetilde{p}+(\delta + \gamma),\theta] \right |,
\end{aligned}
\end{equation}
\end{small}
%where  ̃c and c mean the discrete and continuous position of the ground truth center 
where $\hat{O} \in \mathcal{R}^{H \times W \times 3 }$ is the offset and rotation predicted by BEV head. $\widetilde{c}$ and $c$ mean the discrete and continuous position of the ground truth center. 

\textbf{$z$-axis regression.}
BEV head regresses the $z$-axis position of the target center from the BEV feature map, and predict a map $\hat{Z} \in R^{H \times W \times 1}$,
We compute the error in the $z$-axis center using the $L_1$ loss: 
\begin{small}
\begin{equation}\label{tau_10}
\begin{aligned}
L_{z-axis}=  \left | \hat{Z}_{\widetilde{c}}-Z \right |,
\end{aligned}
\end{equation}
\end{small}
where $\widetilde{c}$ is the discrete object center and $z$ is the ground truth of the $z$-axis center.

\textbf{Segmentation.} 
We learn a MLP to predict a probability $S \in \mathbb{R}^{1024}$ for each point in search region. $s_{j} \in S$ denotes the score whether a point $p_j \in \mathbb{R}^{3}$ is a target point. Specifically, $S$ is constrainted by a standard binary cross entropy loss. The points are considered positive if they are located in the ground truth bounding box, otherwise these points are negative. For enhancing the discriminative features, we concatenate the probability $s_{j}$, coordinate $p_{j}$ and feature $\widetilde{F}_{s,j}^{(3)}$ as $\{s_{j};p_{j};\widetilde{F}_{s,j}^{(3)}\} \in R^{1+3+D}$ for each point in search region. Then enhanced features are feed into Voxel-to-Bev head to predict proposals.

%% file: experiments.tex
\renewcommand{\arraystretch}{1.3} %line height
\section{Experiments}
In this section, we first introduce our implementation details. 
Then we report the results of our method on different benchmark datasets with comparisons to the baseline methods and several state-of-the-art tracking methods. 
It is worth noting that we evaluate the recent advanced trackers in a new way which defines a class-agnostic tracking for 3D SOT. 
Finally, ablation studies are provided to analyze the impact of each component and different design choices.

\subsection{Experimental Settings}  
%\subsubsection{\textbf{Implementation details}\\ } 
\textbf{Datasets.} For 3D single object tracking, we use KITTI~\cite{Geiger2012AreWR} and nuScenes~\cite{Caesar2020nuScenesAM} datasets for training and evaluation. 
Since having no access to the ground truth of the official test set of KITTI dataset, we follow P2B~\cite{Qi2020P2BPN} and use its public training set to train and evaluate different methods. It contains 8 types of objects in $21$ LiDAR sequences, where the scenes $0-16$ are for training, scenes $17-18$ for validation, and scenes $19-20$ for testing. 
For nuScenes dataset, we use its validation set to evaluate the generalization ability of our method. Note that the nuScenes dataset only annotates key frames, so we report the performance evaluated on the key frames.

\textbf{Evaluation metrics.}
%我们使用成功率和精度作为3D单目标跟踪的质量评判指标，其中精度代表预测包围盒与真实包围盒的距离。
Following P2B, we use success and precision ratios as metrics for 3D single object tracking. 
$Success$ calculates the ratio of the intersection over union (IOU) between the prediction and the ground truth bounding box greater than a threshold, 
and $Precision$ measures distance error between center points of the two bounding boxes from 0-2m.

\textbf{Network architecture.} 
We sample $1024$ points in the search region and $512$ points in the template. In the local encoding stage, each point will establish a connection relationship with the points in the 0.3-meter sphere neighborhood. Here we use two layers of GCN for convolution to obtain initialized features of the template and search region. In the feature extraction stage, we set up a three-layers template-aware Transformer module. The output of each layer will be down-sampled by the FPS, and the number of down-sampled points is reduced to $256$ for the first layer, $512$ for the second layer. The network then performs features aggregations in the search region, gradually transferring low-density point cloud features to the highest density. The point-wise feature in the above stages have always maintained $64$ dimensions. Note that in this paper, we adopt the Transformer with multi-head attention having four heads for all experiments.

%\begin{figure*}[ht]
%  \centering
%  \begin{overpic}[width=2.0\columnwidth,tics=10]{visual.pdf}  
%  \end{overpic}
%  \caption{The figure shows the distributions of features visualized by T-SNE\cite{van2008visualizing} method. (a). the visualizations of search features from class-specific testing; (b) the visualizations of search features from class-agnostic testing. For (a)/(b), We show the features distributions of four examples (from top to bottom) from P2B, BAT, V2B, and OST(ours)). The target points are plotted by blue, and the background points plotted by red.}
%  \label{fig:vis}
%  \vspace{-7pt}
%\end{figure*}
\begin{figure*}[h]
	\begin{minipage}{0.49\linewidth}
		\vspace{3pt}
		\centerline{\includegraphics[width=\textwidth]{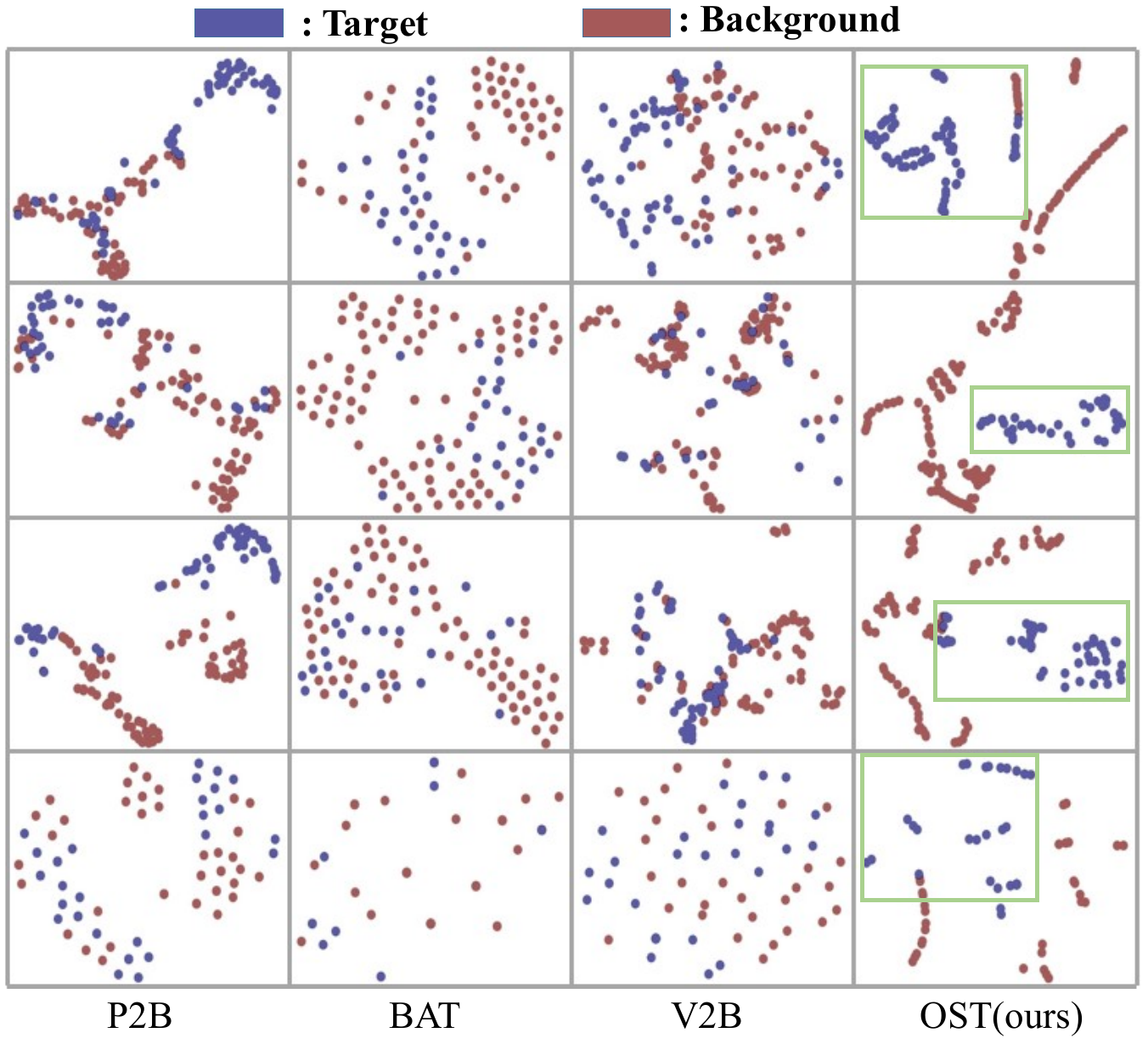}}
		\centerline{(a)}
	\end{minipage}
	\begin{minipage}{0.49\linewidth}
		\vspace{3pt}
		\centerline{\includegraphics[width=\textwidth]{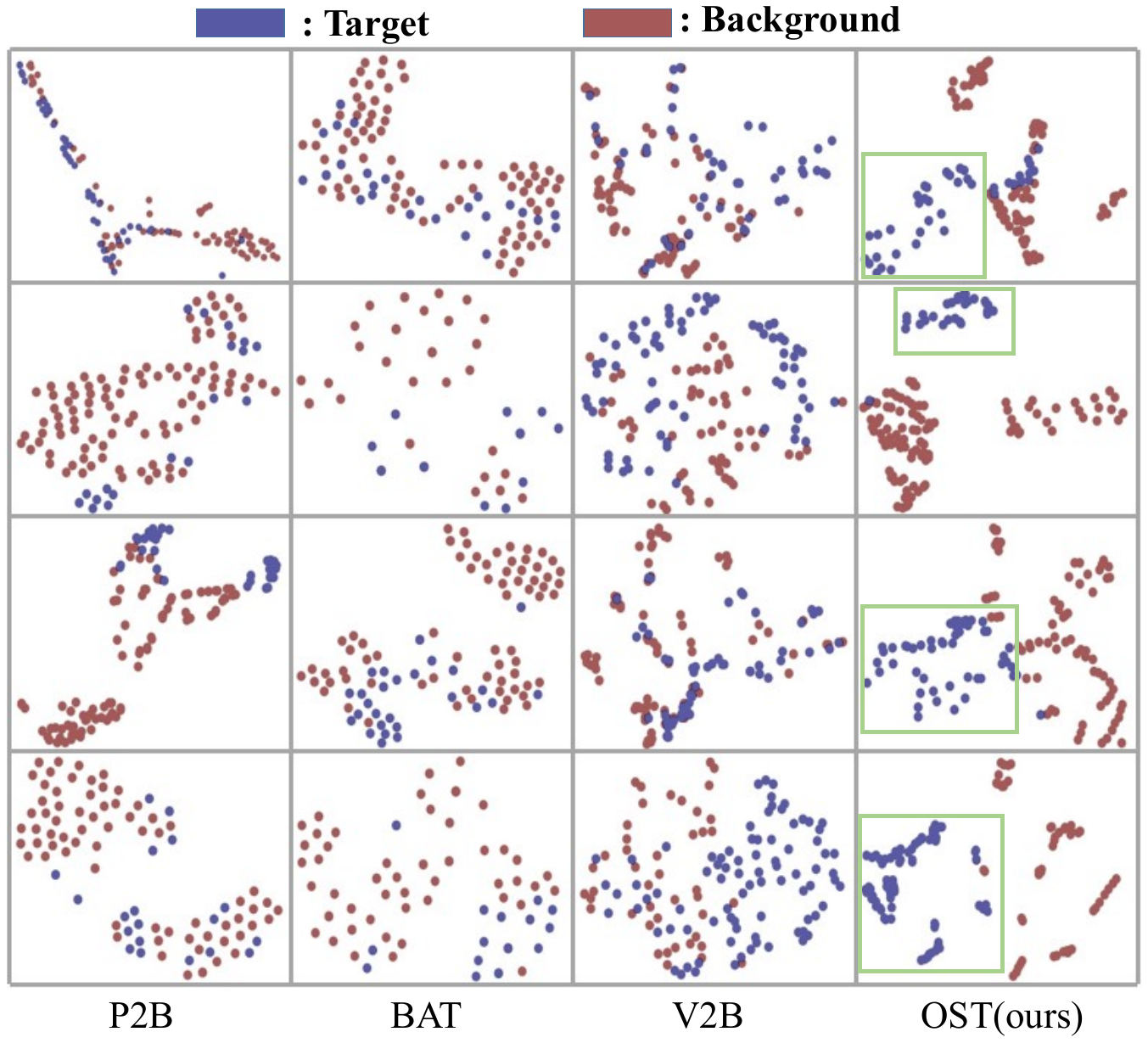}}
		\centerline{(b)}
	\end{minipage}
	\caption{The visualization of the feature distribution of the target and background via T-SNE~\cite{van2008visualizing} method. (a) The visualizations of search features from class-specific testing; (b) The visualizations of search features from class-agnostic testing. (a)/(b) We visualize the search region point cloud features from four different frames in KITTI car dataset. The subfigures in each row are from the same frame, and the subfigures in each column are from the same method. The target points are plotted by blue, and the background points plotted by red.}
	\label{fig:vis}
\end{figure*}

\subsection{Evaluation on class-specific tracking}\label{subsec:class_special}
%\subsection{Comparison with State-of-the-Art Trackers}
%\subsubsection{Evaluation on class-specific tracking} 
%\ 
%\newline 
\subsubsection{Quantitative comparison with State-of-the-Art Trackers} 
%\indent \textbf{Quantitative comparison.}
There has emerged a lot of trackers for 3D SOT task, which incorporate different approaches with Siamese framework. For example, the family of the pointnet++ includes P2B~\cite{Qi2020P2BPN}, BAT~\cite{Zheng2021BoxAwareFE}, MLVSNet~\cite{Wang2021MLVSNetMV}, PTTR~\cite{zhou2022pttr}, and V2B~\cite{Hui20213DSV}; the sparse convolution LTTR; and the Transformer STNet. For comprehensive comparison, we compare our one-stream framework with these sate-of-the-art trackers on the KITTI and nuScenes datasets. To fully demonstrate the tracking performance, we evaluate these methods across four categories per dataset. KITTI covers car, pedestrian, van, and cyclist; while nuScenes is car, pedestrian, truck, and bicycle. It's worth note that all results of nuScenes are tested by models trained in KITTI.

% 介绍实验 对比方法的特点，实验内容
As shown in the Table~\ref{tab:class_special_result}, we summarize and report the performance of these trackers according to the success and precision metrics. Our results are shown in the last row of the table. It can be seen that our method achieves the highest average results across the four categories on KITTI. In particular, compared with the pioneers of the Siamese tracking network, our one-stream method (named OST) significantly outperforms P2B and BAT in all categories. In addition, our OST obtains considerable improvement on car class of KITTI, with 1.5\% and 2.9\% higher than V2B in terms of success and precision. It is also worthy noticing that our OST reaches comparable performance in car class when compared with STNet, even though it equips with a complicated correlation module. Especially, our method is superior to it in pedestrian and van.  
Besides, we use a two layers GCN network to learn point cloud local spatial information in our method, but this module is too simple to efficiently encode the points clouds in other domains. As a result, the results on nuScenes are unsatisfactory for our method. For domain generalization, a powerful and robust spatial encoder module is indispensable. The domain generalization in 3D SOT is our future goal. 

\subsubsection{Qualitative comparison with State-of-the-Art Trackers}
Our method joints the feature extraction and relation modeling by a one-stream backbone while the Siamese framework establish relationship after a Siamese backbone. In order to directly show the discrepancy of above frameworks, we compare the feature distributions of our method (OST) and three representative Siamese frameworks (P2B, BAT, and V2B) in KITTI dataset. Specifically, we use the T-SNE~\cite{van2008visualizing} method to visualize the search region features. For fair comparison, the feature of our OST is visualized after passing both TTM and MFA. As for the Siamese frameworks, we visualize their output features after augmenting by the relational modeling modules. As shown in Figure~\ref{fig:vis}(a), the point features on the target surface and the background are highlighted by blue and red, respectively. From the last column of this figure, we can see that our OST has obvious boundaries, while the features in those Siamese framework (P2B, BAT, and V2B) are mixed together. This convincingly illustrates that our method makes these two types of features distinguishable, which is beneficial to predicting proposals. On the contrary, poor performances make difficultly for those Siamese frameworks to distinguish the target. Benefit from it, our method achieves better results as shown in Table~\ref{tab:class_special_result}.

\begin{figure*}[ht]
  \centering
  \begin{overpic}[width=2.0\columnwidth,tics=10]{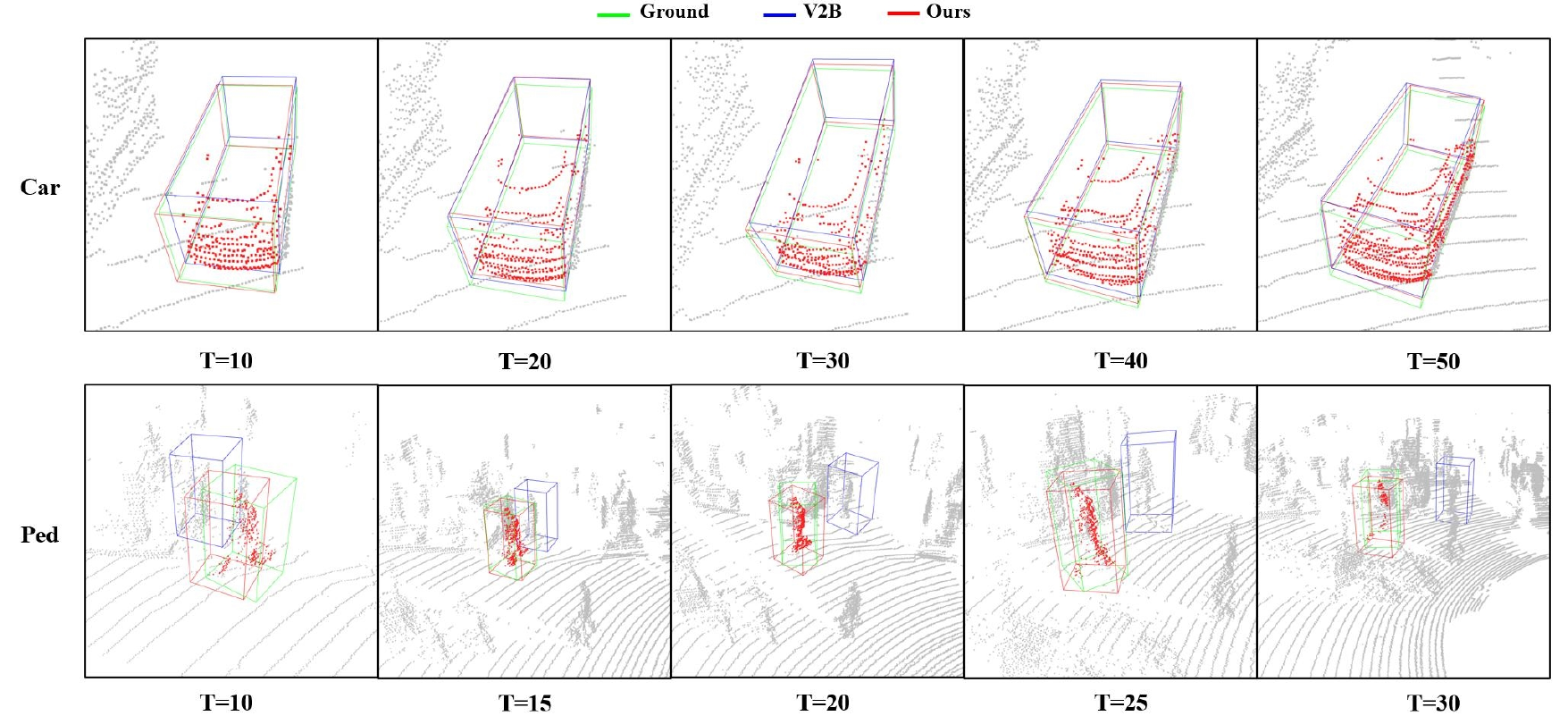}  
  \end{overpic}
  \caption{The Comparison with V2B on KITTI dataset. The first row shows the results of the same scene's different frames in car category and the second row shows the results in pedestrian (ped) category. The visualization results illustrate our method has stronger immunity to interference of similar objects than v2b.}
  \label{fig_results:class_special}
  \vspace{-7pt}
\end{figure*}

In Figure~\ref{fig_results:class_special}, we visualize the results of our method (red boxes) and V2B (blue boxes). The first row shows the results of the same scene's different frames in car category and the second row shows the results in pedestrian category. We can see that our method can still identify the target even in the pedestrian dataset full of interference. In contrast, in the case of interference from similar objects, V2B cannot capture the target. The visualization results illustrate our method has stronger immunity to interference of similar objects than V2B.

\begin{table}[htb]
  \caption{Settings for Class-agnostic Tracking in Point Clouds.}
  \label{tab:class_agnostic_setting}
  \begin{center}
  \begin{tabular}{l|l|cc}
    \toprule
     \multirow{3}{*}{Split} & \multirow{3}{*}{Category} &\multicolumn{2}{c}{KITTI} \\
     \cmidrule(llr){3-4} 
       &    & setting-1  & setting-2   \\
        \midrule 
   \multirow{4}{*}{Train}     &  & Pedestrain  & Car    \\
         &                       & Van  & Van           \\
         & -                     & Cyclist  & Cyclist   \\
         &                       & (8123)  & (23045)    \\  % 19522 4600 1994 1529
    \midrule 
   \multirow{6}{*}{Test}     &  \multirow{4}{*}{Observed}           & Pedestrain    & Car    \\
                             &                                      & Van           & Van    \\
                             &                                      & Cyclist       & Cyclist   \\
                             &                                      & (7644)        & (7980)   \\
    \cmidrule(llr){2-4}
                            &\multirow{2}{*}{Unseen}                & Car  & Pedestrain  \\
                            &                                       &(6424)  & (6088)   \\
    \bottomrule
  \end{tabular}
  \end{center}
\end{table}

\subsection{Evaluation on class-agnostic tracking}

\subsubsection{Settings for Class-agnostic Tracking in Point Clouds}
At this stage, the advanced methods of 3D SOT invariably follow a paradigm that the model is trained in a specific class and tested in this class. However, this mode requires that the class to which the tracked object belongs is present in the training dataset. This is strict hypothesis to achieve in reality, on the contrary, dealing with challenging and out-of-distribution (OOD) scene in tracking task requires the model to be able to track any objects. Therefore, we believe that models of advanced single object tracking methods should not only perform well on a specific class, but also achieve better results on unseen or OOD classes.

%\begin{table}[t]
%  \caption{Settings for Class-agnostic Tracking in Point Clouds.}
%  \label{tab:class_agnostic_setting}
%  \begin{center}
%  \begin{tabular}{lcc|cc}
%    \toprule
%    \multirow{3}{*}{Category} &\multicolumn{2}{c}{KITTI} & \multicolumn{2}{c}{nuScenes}\\
%    \cmidrule(lr){2-3} \cmidrule(lr){4-5}
%              & setting-1  & setting-2   & Setting-1 & Setting-2 \\
%    \midrule 
%          & Pedestrain  & Car   & Pedestrain  & Car \\
%    Observed & Van  & Van   & Truck & Truck \\
%    Category & Cyclist  & Cyclist   & Bicycle & Bicycle \\
%          & (7644)  & (7980)   & (49106) & (80038) \\     
%    \midrule  
%         Unseen & Car  & Pedestrain   & Car & Pedestrain \\
%        Category   &(6424)  & (6088)   & (64159) & (33227) \\
%    \bottomrule
%  \end{tabular}
%  \end{center}
%\end{table}

%In order to verify the effectiveness of our method, following \cite{tian2022towards}, we set up experiment as shown in Table.\ref{tab:class_agnostic_setting}. 
To reflect the advantages of the proposed one-stream method on 3D SOT, we use a class-agnostic setup to evaluate the different methods according to the literature~\cite{Tian2022TowardsCT}. As shown in the Table~\ref{tab:class_agnostic_setting}, it provides two settings on KITTI dataset. %and nuScenes dataset. 
%we set up the KITTI dataset and the nuScenes dataset respectively. 
As for the KITTI dataset, the categories commonly used for tracking are cars, pedestrians, vans, and cyclists, and We divide the four categories into two groups by two ways. In setting-1, we divide pedestrians, vans, and cyclists as the group for training, and cars for testing. In setting-2, we divide cars, vans, and cyclists as the group for training, and pedestrians for testing. The categories used for training are the ones observed by the model and the others are the unseen. Under such experimental settings, on the one hand, we can conduct class-agnostic experiments to verify the effectiveness of our method when test in unseen category, on the other hand, the robustness of the model can be measured when tested on the observed category.

\subsubsection{Results of Class-agnostic Tracking}
We conduct class-agnostic experimental comparisons with current state-of-the-art object tracking methods, including quantitative experimental comparisons and qualitative experimental comparisons.
% 类无关 量化
\begin{table}[t]
  \caption{Performance comparison with advanced methods in class-agnostic experiment. The Success/Precision are reported and the best results are highlighted in bold.}
  \label{tab:class_agnostic_results}
  \centering
  \setlength{\tabcolsep}{1.5mm}{
  \begin{tabular}{l|cccc}
    \toprule
    \multirow{4}{*}{Methods}  & \multicolumn{4}{c}{KITTI}   \\
    \cmidrule(lr){2-5} 
              & Setting-1         & Setting-1    & Setting-2          &Setting-2    \\
              & Observed &Unseen & Observed & Unseen \\
    \midrule  
     P2B	&38.6/57.4	&32.4/41.0	&55.6/70.1	&24.2/45.4 \\
     BAT	&33.9/48.9	&24.9/33.7	&59.6/73.1	&12.9/21.8 \\
     V2B    & \textbf{52.1/74.7} & 47.8/59.6 & 66.1/77.1 & 24.4/43.7   \\
     %OST(ours)   & 48.1/73.4 & 50.2/61.2 & 69.7/81.4 &  29.5/53.3
     OST(ours) & 50.0/74.2 & \textbf{49.2/62.4} & \textbf{68.1/79.7} & \textbf{28.9/56.1} \\
%    \bottomrule
%    \multirow{4}{*}{Methods}    & \multicolumn{4}{c}{nuScenes} \\
%    \cmidrule(lr){2-5} 
%                & Setting-1         & Setting-1    & Setting-2   &Setting-2 \\
%               & Observed & Unseen & Observed & Unseen \\
%    \midrule  
%     P2B	 &31.1/46.1 &34.3/35.3 &42.5/45.9&14.6/33.7\\
%     BAT	 & 30.9/44.0 & 33.6/35.4 & 38.6/40.5  &16.9/37.1\\
%     V2B   &  &  &  &     \\
%     OST(Ours)   &  &  &   &    \\
    \bottomrule
  \end{tabular}}
\end{table}

\textbf{Quantitative comparison.}
% 方法介绍
The state-of-the-art Siamese trackers (P2B~\cite{Qi2020P2BPN}, BAT~\cite{Zheng2021BoxAwareFE}, V2B\cite{Hui20213DSV}) extract features base on a parameter-shared backbone, then establish relationship between template and search region by independent relational modeling module. Our OST joints feature extraction and relation modeling by the TTM. For fully comparing with those advance methods, We report the results in class-agnostic experiment on the KITTI. As shown in Table~\ref{tab:class_agnostic_results}, we summarize and report the performance of these trackers according to the success and precision metrics. Our results are shown in the last row of each table part. As we can see, compared with BAT and P2B, our method achieves a great improvement in observed categories. Furthermore, compared with the strong baseline V2B, our method also achieves competitive results. More importantly, it is obvious that our method has achieved more prominent progress in unseen categories with a large improvement in success and precision. In detail, the proposed method improves 1.4\%/2.8\% under setting-1 and 4.5\%/12.4\% under the setting-2 relative to V2B. 

These results on the unseen category benefit from our proposed template-aware feature learning. In the feature learning stage, the search region extract features according to the template, and the segmentation network divides the search region into foreground and background based on the input template, not the category. While these Siamese methods based on PointNet++ learn the class-specific shape information. The class priors is positive for tracking the observed classes, but negative for unseen classes. 

 %Making the network fully aware of template information as much as possible rather than a prior for a given class is the key to solving class-agnostic tracking. Only in this way can the tracker give accurate results in the face of agnostic categories.

% 类无关特征图
%\begin{figure}[htb]
%  \begin{center}
%    \includegraphics[width=0.9\linewidth]{picture/class_special_feature.pdf}
%  \end{center}
%  \setlength{\abovecaptionskip}{-0.15cm}
%  \setlength{\belowcaptionskip}{-0.cm}
%  \caption{The figure shows the distribution of the features extracted by different methods from class-agnostic testing. We show the feature distribution of four examples (from top to bottom in the figure) in different methods(from left to right are P2B, BAT, V2B, Ours).}
%  \label{fig_fea:class_agnostic}
%\end{figure}

\textbf{Qualitative comparison.}
In the quantitative experiments above, our method is able to outperform other methods on the unseen categories because our template-aware feature extraction comes into play. To verify this effect, we visualize the intermediate features of the network extracted from the testing phrase via T-SNE~\cite{van2008visualizing}. For the Siamese-based approaches, P2B, BAT and V2B, we visualize the features after the relation modeling. While for our approach, we directly visualize the features mentioned in the one-stream backbone. The Figure~\ref{fig:vis}(b) shows the feature visualization of unseen category based on the setting-1. The visualization results of our method are on the far right. As we can see, for categories that have not been seen in training time, our feature extractor can also effectively distinguish target points from background points, while P2B, BAT and V2B are not well differentiated. Our OST can effectively distinguish those target points even though it is exposed to the object absent in the training phrase, which it means our method get the target semantics rather than category priors. This also explains the excellent performance of our quantitative experiments in the previous section.

\begin{figure*}[t!]
  \centering
  \begin{overpic}[width=2.0\columnwidth,tics=10]{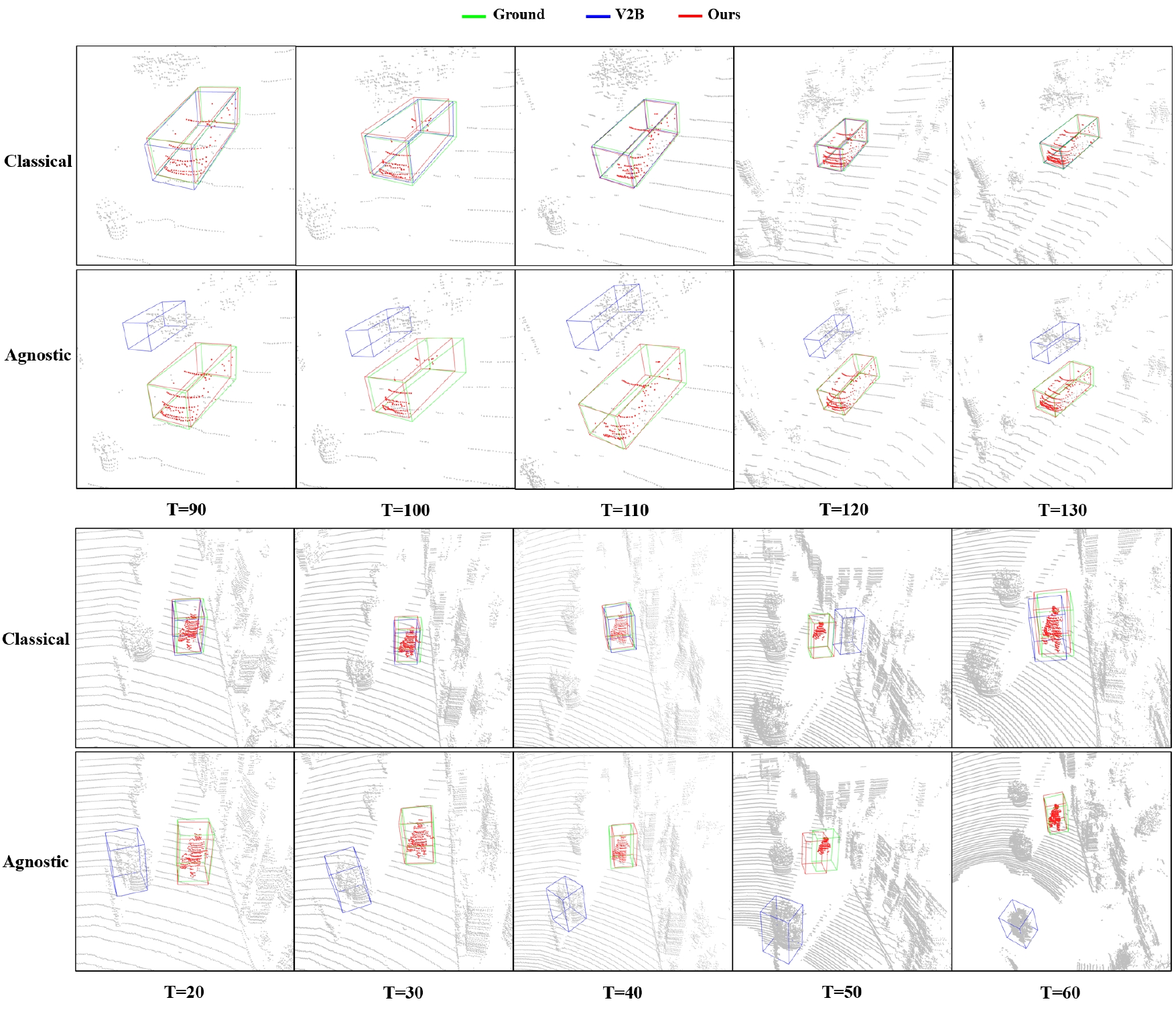}
  \end{overpic}
  \caption{Visualization results to demonstrate our methods have powerful generalization ability. "Classical" and "Agnostic" means class-specific setting and class-agnostic setting separately. Although V2B could capture the target under normal settings, it completely lost tracking ability when the category of the tracking target changed.}
  \label{fig_results:class_agnostic}
  \vspace{-7pt}
\end{figure*}

We further compare the results from class-specific and class-agnostic experiment at same scene. In Figure~\ref{fig_results:class_agnostic}, "Classical" and "Agnostic" means class-specific setting and class-agnostic setting separately. For class-specific, tracking objects category is exposed during training while it is not exposed for class-agnostic. As shown in the Figure~\ref{fig_results:class_agnostic}, we can see that our method can capture the target regardless of whether it has been exposed during training. In contrast, it fails if V2B has not seen such objects while training.

\begin{figure}[htb]
  \begin{center}
     \includegraphics[width=1\linewidth]{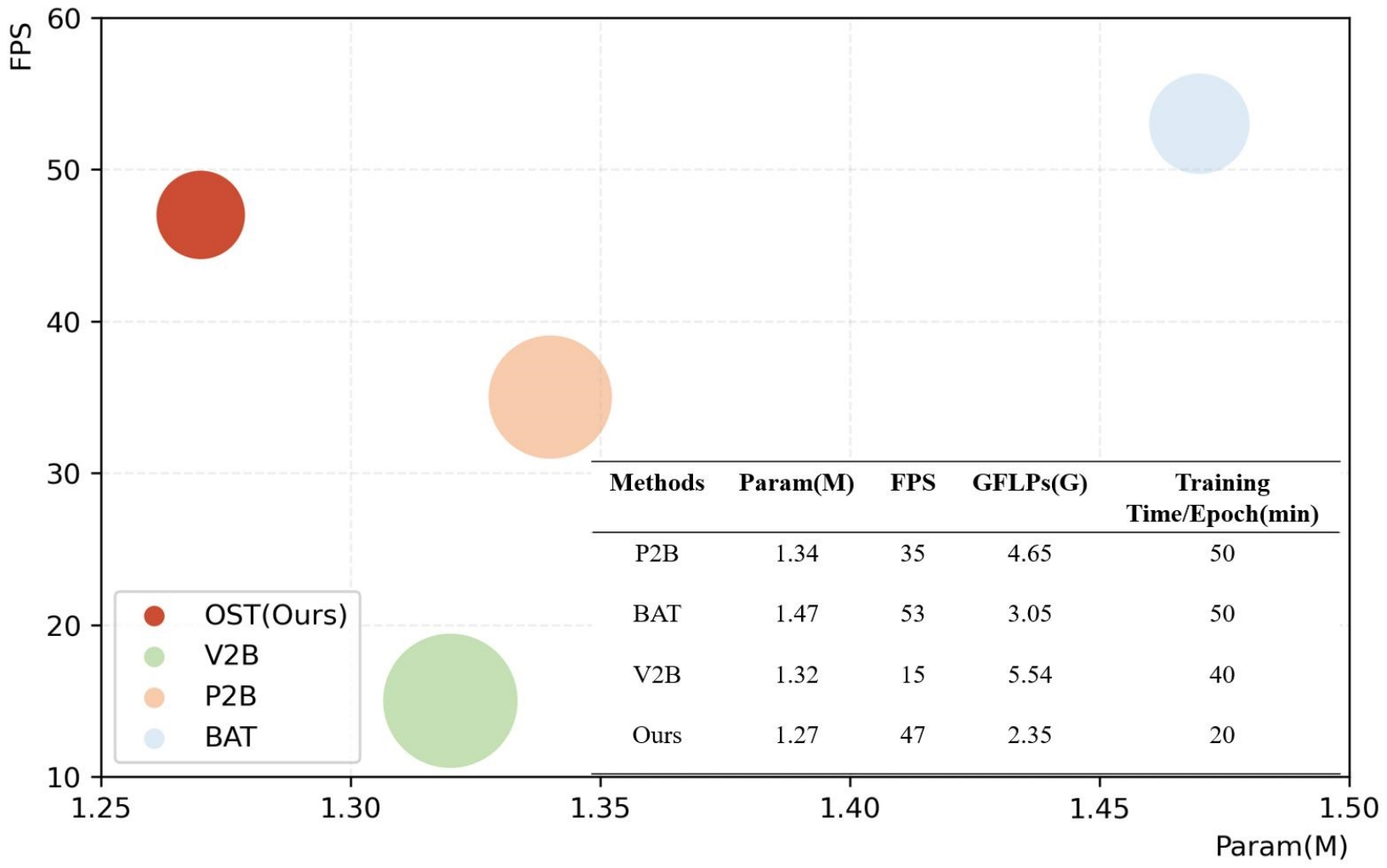}
  \end{center}
  \setlength{\abovecaptionskip}{-0.15cm}
  \setlength{\belowcaptionskip}{-0.cm}
  \caption{The Comparison from parameter quantity, floating-point operations per second (FLOPs), frames per second (FPS), and training time per epoch of different methods. Horizontal and vertical coordinates denotes the parameter quantity and FPS respectively and point size denotes FLOPs. The method which is closer to the top left corner with smaller size is better.}
  \label{fig:cost}
\end{figure}

\subsection{The computational cost of different methods}
%\begin{table}[h]
%  \setlength{\abovecaptionskip}{-0cm}
%  \setlength{\belowcaptionskip}{-0cm}
%  \caption{The computational cost of different methods.}
%  \label{tab:computate_cost}
%  \center
%  \begin{tabular}{lcccc}
%    \toprule
%    Method & Param & FLOPs &FPS & Training time     \\
%    \midrule
%    P2B & 1.34M & 4.28G & 17 & $\sim$ 13h \\
%    BAT & 1.47M & 5.53G & 18 & $\sim$ 8h  \\
%    V2B & 1.35M & 5.47G & 13 & $\sim$ 8h  \\
%    Ours & 1.22M & 2.32G & 30 & $\sim$ 4h \\
%    \bottomrule
%  \end{tabular}
%\end{table}

We compare the proposed OST with P2B~\cite{Qi2020P2BPN}, BAT~\cite{Zheng2021BoxAwareFE} and V2B~\cite{Hui20213DSV} in terms of parameters, FLOPs, FPS, and training time. Specifically, we evaluate our model on the car category in the KITTI dataset and all experiment is on a single TITAN RTX GPU. For training time, we calculate the average time per epoch with batchsize 24. As shown in Figure~\ref{fig:cost}, compared with other methods, we can see that our method has fewer parameters and less FLOPs, and the training time is reduced exponentially. Especially, it is very fast at the testing phase, the FPS can reach 47it/s, and it is noting that only 12.3 ms for network forward propagation. Compared with Siamese backbones (P2B, BAT, and V2B), our template-aware feature extraction backbone can help us complete feature extraction and relationship association at one time, which is makes our method not need a separate relation modeling module to generate relationship, so our method is more lightweight and faster.

\subsection{Ablation study}\label{ablation}

\subsubsection{The effect of different components}
\begin{table}[htb]
  \setlength{\tabcolsep}{0.8mm}
  \setlength{\abovecaptionskip}{-0.cm}
  \setlength{\belowcaptionskip}{-0.cm}
  \caption{The ablation study results of different components on car category.}
  \label{tab:self-ablation}
  \center
  \begin{tabular}{ccc|cccc}
    \toprule
   Local    & TTM   & MFA & Car & Pedestrian & Van & Cyclist \\
    \midrule
    $\surd$ & $\quad$ & $\quad$  & 68.3/80.5  & 47.4/76.2 & 55.0/65.6 & 44.4/54.4 \\
    $\surd$ & $\surd$ & $\quad$  & 71.7/83.5  & 48.9/76.7 & 56.4/66.4 &  49.2/60.4 \\
    $\surd$ & $\surd$ & $\surd$  & 72.0/84.2  & 51.4/82.6 & 57.5/68.2 &  49.2/60.4 \\
    \bottomrule
  \end{tabular}
  \end{table}

To justify the effectiveness of each component, we evaluate these components across four categories in KITTI dataset covers cars, pedestrians, vans, and cyclists. We report the success and precision in Table~\ref{tab:self-ablation}. We set the local encoding module named local, the template-aware Transformer module named TTM and the multi-scale feature aggregation module named MFA in the table. It can be seen that when adding the TTM improves the tracking performance by 3.4\% and 3.0\% in car categories, and adding the MFA makes the performance continue to increase by 0.3\% and 0.7\%. These verify that the network module we have set up is valid. The Transformer layer performs feature extraction and relationship modeling according to template during feature extraction, so that the points in the search region have the information from template. Multi-scale feature aggregation methods can perceive features at different layers to compensate for the spatial information lost when the network layer is deepened. As a result, adding these module makes the performance of our method improve.

\subsubsection{The effect of the number of TTM}

\begin{figure}[htb]
  \begin{center}
     \includegraphics[width=1\linewidth]{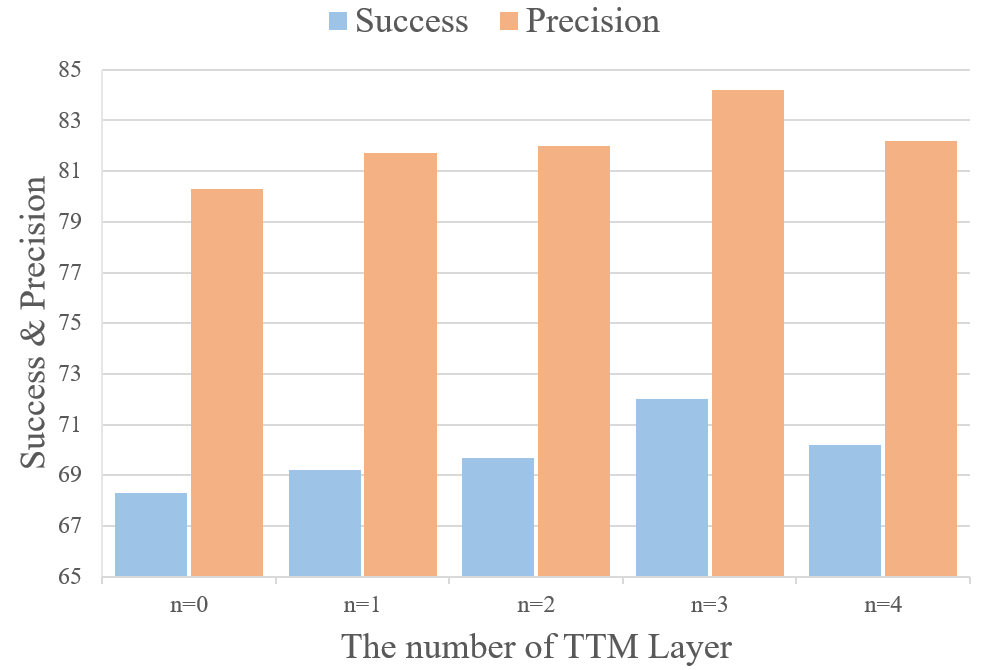}
  \end{center}
  \setlength{\abovecaptionskip}{-0.15cm}
  \setlength{\belowcaptionskip}{-0.cm}
  \caption{Ablation study to explore the number of TTM's influence on the method. The figure shows the results of our method which set by different number (n=0,1,2,3,4) of TMM layers in the KITTI car dataset.}
  \label{fig:number_layer}
\end{figure}

The number $n$ of the template-aware Transformer module layers is a key parameter for feature extracting. Here we study the effects of different values of $n$ on tracking accuracy. Figure~\ref{fig:number_layer} shows the network performance provided by different numbers of layers including $0$, $1$ , $2$ , $3$ , and $4$, respectively. In the figure, the horizontal axis represents the number of layers, and the vertical axis represents the tracking accuracy. 
Then the green line shows the success our method performed on car category in KITTI dataset and the blue line represent the  precision. As shown in the Figure~\ref{fig:number_layer}, we can see that our tracker achieves the best performance when $n=3$ with success 72.0\% and precision 84.2\%.

\subsubsection {Comparison of different multi-scale aggregation strategies}

%\begin{figure}[t]
%  \begin{center}
%     \includegraphics[width=1\linewidth]{picture/class_agnostic_Siamese_onestream.pdf}
%  \end{center}
%  \setlength{\abovecaptionskip}{-0.15cm}
%  \setlength{\belowcaptionskip}{-0.cm}
%  \caption{The figure shows the results tested in unseen category(Car) from KITTI dataset when doing class-agnostic setting-1 experiments. }
%  \label{fig:siamese_onestream_view}
%\end{figure}

%\textbf{Completion and segmentation ---\textbf{$Table$}}

\begin{table}[ht]
  \setlength{\abovecaptionskip}{-0.cm}
  \setlength{\belowcaptionskip}{-0.cm}
  \caption{Ablation study in the KITTI dataset. The comparison with feature aggregation designed with different methods.}
  \label{tab:multi-direction}
  \center
  \begin{tabular}{l|cccc}
    \toprule
    Method & Car & Pedestrian  & Van & Cyclist   \\
    \midrule
      OST/w.usual    & 70.9/82.9 & 44.2/75.6  &   52.3/62.0	& 47.9/60.0  \\
      OST/w.specific  & 72.0/84.2 & 51.4/82.6  &   57.5/68.2  & 49.2/60.4  \\
    \bottomrule
  \end{tabular}
\end{table}

%As shown the figure.\ref{fig:transformer&multi-scale}, our feature aggregation module figure.\ref{fig:transformer&multi-scale}(c) is different from common module figure.\ref{fig:transformer&multi-scale}(b). 
For feature aggregating, the usual practice is that spread information from deep layer to shallow layer as Figure~\ref{fig:transformer&multi-scale}(b). 
In our method, we aim to perceive features at different layers to compensate for the spatial information lost when the network layer is deepened. 
With the idea, we proposed the tracking-specific method that spread information from shallow layer to deep layer. We compare the tracker performance by our setting with common setting. In Table~\ref{tab:multi-direction}, we report the results of two setting, the usual practice named OST/w.usual and our method named OST/w.specific. We can observe in the table that our method (OST/w.specific) achieves the better results across the four categories on KITTI. 
In detail, our method improves by 1.1\%/1.3\%, 7.2\%/7.0\%, 5.2\%/6.2\%, 1.3\%/0.4\% on the four categories, respectively. 

\begin{table}[ht]
  \setlength{\abovecaptionskip}{-0.cm}
  \setlength{\belowcaptionskip}{-0.cm}
  \caption{Ablation study in the KITTI dataset. The comparison with Siamese networks designed with different methods.}
  \label{tab:mainstream_onestream}
  \center
  \scalebox{0.9}{
    \begin{tabular}{l|cccc}
    \toprule
    Method & Car & Pedestrian  & Van & Cyclist   \\
    \midrule
    Siamese/w.P2B-xcorr   & 71.4/83.0 &  48.3/74.7	&  	58.7/69.0  & 40.3/54.5  \\
    Siamese/w.V2B-xcorr   & 71.9/83.2 &	 49.2/77.4	&   57.4/64.8  & 40.3/49.0  \\
    One-stream            & 72.0/84.2 &  51.4/82.6  &   57.5/68.2  & 49.2/60.4  \\
    \bottomrule
  \end{tabular}
  }
\end{table}

\subsubsection{Comparison of Siamese network and one-stream network}
In addition to the comparison with state-of-the-art Siamese-based trackers, we also disassemble our one-stream framework into Siamese structures for experimental comparison. Specifically, the Transformer module is performed separately instead of splicing the template with the search region, and both branches perform multi-scale feature enhancement. Then we associate the two sides using the relational modeling module in P2B and V2B. The results of the above steps are used to predict proposals by the way in our method including segmentation and Voxel-to-Bev head. The experimental results based on the above settings are shown in Table~\ref{tab:mainstream_onestream}. The Siamese framework with P2B relation modeling module written as Siamese/w.P2B-xcorr and the Siamese framework with BAT relation modeling module written as Siamese/w.BAT-xcorr. Although Siamese networks have a powerful relational modeling module and their computational complexity is much higher, our OST can also achieves slightly better performance compared with them, especially in cyclist. 

%孪生网络与单分支
\subsubsection{Comparison of segmentation and completion}
 
%分割 补全
\begin{table}[ht]
  \setlength{\abovecaptionskip}{-0.1cm}
  \setlength{\belowcaptionskip}{-0.1cm}
  \caption{Ablation study using the KITTI dataset. The results on different module for augmenting features are reported.}
  \label{tab:completing-2-segmentation}
  \center
  \begin{tabular}{l|cccc}
    \toprule
    Method &   Car & Pedestrian  & Van & Cyclist      \\
    \midrule
    With completion   & 70.0/81.9  &50.1/79.0  &53.0/62.8  & 40.7/50.3\\
    With segmentation   & 72.0/84.2 &  51.4/82.6  &   57.5/68.2  & 49.2/60.4  \\
    \bottomrule
  \end{tabular}
\end{table}

In our method, we use a 3D detection head provided by V2B~\cite{Hui20213DSV} for proposal. Differently, we enhance the features for proposal in another way.
V2B uses the completion method to complete the points in the search region and uses the loss to constrain the feature learning in the search region. They aim to enhance the shape information of the points features. Here we use the segmentation method to enhance the features. The purpose is to distinguish the foreground and background points of the search region. 
In Table~\ref{tab:completing-2-segmentation}, we give a comparison of completion and segmentation, and we report the results on KITTI measured by success and precision. As shown in the Table~\ref{tab:completing-2-segmentation}, the segmentation method performs better than the completion method on all four categories. In detail, Our method improves by 2.0\%/2.3\%, 1.3\%/3.6\%, 4.5\%/5.4\%, 8.5\%/10.1\% on the four categories, respectively. 

% 可视化结果 special

%% file: conclution.tex
\section{Conclusion and Discussion}

In this paper, we proposed a One-stream framework for 3D single object tracking, including two core parts, a template-aware Transformer module (TTM) and a multi-scale feature aggregation module (MFA). 
The former realizes a novelty of integrating the relation modeling into feature extraction, while the latter reaches a strong alliance between the spatial and semantic information.
In this way, our method can efficiently resolve arbitrarily specified targets while saving a lot of computational cost.
On the one hand, we demonstrated through qualitative and quantitative experiments that our method can achieve competitive performance in traditional class-specific tracking task. On the other hand, in a new setup that raises a class-agnostic 3D SOT, our method also brings considerable improvement compared to the recent advanced trackers.
In the future, considering class-agnostic object tracking frequently occur in practical applications, we will carry on discussing and exploring this problem further from the perspective of general shape encoder and adaptive relation modeling.